\documentclass[a4paper, 10pt, conference]{ieeeconf}      % Use this line for a4 paper

\IEEEoverridecommandlockouts       % This command is only needed if 
\usepackage{color}
\usepackage{float}
\usepackage{makecell}
\usepackage{longtable}
\usepackage{stfloats}
\usepackage{multirow}
\usepackage{amsmath}
\usepackage{graphicx}
\usepackage{bm}
\usepackage{subfigure}
\usepackage{threeparttable}
\usepackage{balance}

\title{\LARGE \bf
Contrastive Learning for Automotive mmWave Radar Detection Points Based Instance Segmentation
}

\author{Weiyi Xiong$^{1*}$, Jianan Liu$^{2*}$, Yuxuan Xia$^{3}$, Tao Huang$^{4}$, Bing Zhu$^{1\dag}$ and Wei Xiang$^{5}$
\thanks{$^{1}$W.~Xiong and B.~Zhu are with School of Automation Science and Electrical Engineering,
        Beihang University, Beijing 100191, P.R.~China 
        {\tt\small weiyixiong@buaa.edu.cn} (W.Xiong); 
        {\tt\small zhubing@buaa.edu.cn} (B.Zhu)}
\thanks{$^{2}$J.~Liu is with Vitalent Consulting,
        Gothenburg 41761, Sweden, and Silo AI, Stockholm, Sweden 
        {\tt\small jianan.liu@vitalent.se, jianan.liu@silo.ai}}
\thanks{$^{3}$Y.~Xia is with Department of Electrical Engineering, 
        Chalmers University of Technology, Gothenburg 41296, Sweden
        {\tt\small yuxuan.xia@chalmers.se}}
\thanks{$^{4}$T.~Huang is with College of Science and Engineering,
        James Cook University, Cairns, Australia
        {\tt\small tao.huang1@jcu.edu.au}}
\thanks{$^{5}$W.~Xiang is with School of Engineering \& Mathematical Sciences,
        La Trobe University, Melbourne, Australia
        {\tt\small w.xiang@latrobe.edu.au}}
\thanks{$^{*}$Both authors contribute equally to the work and are co-first authors.}
\thanks{$^{\dag}$Corresponding author.}
\thanks{This paper has been accepted by 2022 IEEE 25th International Conference on Intelligent Transportation Systems (ITSC). Digital Object Identifier 10.1109/ITSC55140.2022.9922540}
}

\begin{document}

\maketitle
\thispagestyle{empty}
\pagestyle{empty}

%%%%%%%%%%%%%%%%%%%%%%%%%%%%%%%%%%%%%%%%%%%%%%%%%%%%%%%%%%%%%%%%%%%%%%%%%%%%%%%%
\begin{abstract}
The automotive mmWave radar plays a key role in advanced driver assistance systems (ADAS) and autonomous driving. 
%nowadays since it is a low-cost sensor that can work reliably under all weather conditions. 
%By using sparse detection points provided by the automotive radar, instance segmentation can provide approximated instances with semantic information, resulting in a better comprehension of road traffic and more reliable input to the tracker for tracking multiple objects. 
Deep learning-based instance segmentation enables real-time object identification from the radar detection points.
% and precise annotation is required for training the algorithm.
In the conventional training process, accurate annotation is the key. However, high-quality annotations of radar detection points are challenging to achieve due to their ambiguity and sparsity.
%More advanced instance segmentation techniques have also been developed as the raise of deep learning, and these techniques necessitate precise annotation. 
To address this issue, we propose a contrastive learning approach for implementing radar detection points-based instance segmentation.
We define the positive and negative samples according to the ground-truth label, apply the contrastive loss to train the model first, and then perform fine-tuning for the following downstream task. In addition, these two steps can be merged into one, and pseudo labels can be generated for the unlabeled data to improve the performance further. Thus, there are four different training settings for our method.
Experiments show that when the ground-truth information is only available for a small proportion of the training data, our method still achieves a comparable performance to the approach trained in a supervised manner with 100\% ground-truth information. 
%As a result, our suggested contrastive learning strategy could be potentially used in the real-world automotive radar detection points based instance segmentation challenge, even when ground-truth information is only partially available.

\end{abstract}

%%%%%%%%%%%%%%%%%%%%%%%%%%%%%%%%%%%%%%%%%%%%%%%%%%%%%%%%%%%%%%%%%%%%%%%%%%%%%%%%
\vspace{-1mm}
\section{INTRODUCTION} \label{sec introduction}

Automative mmWave radars, LiDARs and cameras are all important sensors for autonomous driving. A camera can take images that contain rich information such as the colors and edges of objects, and a dual camera system can also determine the distance; the LiDAR emits laser waves and receives them after reflection, generating dense LiDAR points that distribute on the surface of objects so that the shape of each object is obtained. Although cameras and LiDARs provide data which are easy to understand by both humans and computers, there are some drawbacks of them. 
Firstly, they are not able to work in certain weather conditions, such as heavy rain, snow or fog \cite{radar}. In addition, some objects cannot be detected in the presence of occlusion. Last but not least, they do not provide the velocity information, which is important for autonomous driving. As a result, mmWave radars are indispensable in the Advanced Driver Assistance Systems (ADAS), as they can work in all-weather conditions, detect occluded objects, and measure the radial velocity of each object \cite{radar_survey}.

%However, the sparsity of detection points results in the difficulty for scene understanding from radar data. 
However, the sparsity of radar detection points makes it challenging for scene understanding.
Modern deep learning methods could be employed to solve this problem, but those methods need massive labeled data to achieve high performance \cite{radar_ins_seg_1}\cite{radar_ins_seg_2}. Typically, the radar data are manually annotated by human experts \cite{radarscenes} or semi-automatically annotated by certain approaches for simplification \cite{CARRADA}\cite{CRUW}.
%But labeled data from cameras or LiDARs are usually required in either method to guarantee high accuracy, which is costly and hard to balance the time spent and annotation quality. 
But labeled data from cameras or LiDARs are usually required in either method to guarantee high accuracy, which is costly.
Moreover, the radar detection points are often semantically ambiguous. Thus, it is difficult to annotate them with a semantic or instance label. To solve the aforementioned issues, we propose a contrastive learning-based method that only relies on a few labeled training data to perform instance segmentation on radar detection points.

Contrastive learning is a learning strategy that aims to train the model in a self-supervised fashion by finding a proper representation of unlabeled data. For each datum (also called an anchor datum), data which have high similarity with it are defined as its positive samples, while those dissimilar to it are defined as its negative samples. 
As the first step of contrastive learning, properly defining positive and negative samples are important to let the model learn to extract corresponding features. This process is important as it can help in performing the downstream task efficiently.
%By designing a proper way to define positive and negative samples (note that labels are not necessarily required to define them), the model can learn to extract features which are helpful for solving the downstream task efficiently, as the first step of the contrastive learning.
In this step, the contrastive loss and the backpropagation process tries to pull apart the features of negative samples learned by the model, and draw the features of positive samples closer at the same time. 
%To solve the downstream task, a downstream task fine-tuning step is followed.
To solve the downstream task, a fine-tuning step is followed.
In this step, a few labeled data are taken as the input of the model, the parameters of the backbone are frozen, and the model is trained in a supervised learning strategy.

In our work, we follow \cite{radar_ins_seg_1} to apply the semantic segmentation-based clustering method for radar points instance segmentation, while adopting the contrastive learning strategy in replace of the traditional supervised learning strategy to train the semantic segmentation model. Specifically, our contributions are threefold:
\begin{itemize}
\item According to the best of authors' knowledge, we are the first to adopt contrastive learning on the radar detection points-based instance segmentation task to tackle the issue of insufficient point-wise annotation of radar detection points. We expect to inspire more investigations on radar-based perception without sufficient labeled radar data through this study.
\item %We propose an efficient contrastive learning-based strategy for semantic segmentation with radar detection points, and thus improving the performance of radar detection points-based instance segmentation by applying the semantic segmentation-based clustering, under the constraint of limited labeled data. 
We propose an efficient contrastive learning-based strategy for semantic segmentation with radar detection points. By training the model in the proposed contrastive learning strategy, the performance can reach a satisfying level under the constraint of limited labeled data.
Several settings can be applied to our proposed strategy, i.e., fully-supervised setting/non-joint training, semi-supervised setting/non-joint training, fully-supervised setting/joint training and semi-supervised setting/joint training.
%In the fully-supervised training, only the small number of labeled training data are taken as input, while the massive unlabeled training data are also employed in the semi-supervised setting. The representation learning and downstream task fine-tuning are trained in sequence for non-joint training, and the joint training learns the representations of points and the downstream task at the same time. By choosing appropriate training settings, the training time and the performance can be balanced.
\item Experiments show that the model trained with the proposed contrastive learning strategy in all settings outperforms that trained in a supervised learning manner when only a small proportion (5\%) of labeled training data are available. The performance of the latter is 75.59\% mean coverage (mCov) and  70.86\% mean average precision with the IoU threshold of 0.5 (mAP$_{0.5}$), while the different settings of our method obtain an improvement of about 2\%-3.5\% without introducing any additional inference time and memory consumption.
\end{itemize}

The rest of the paper is organized as follows. Section \ref{sec related} introduces the related work, including LiDAR and radar based instance segmentation methods and contrastive learning strategies in the field of computer vision. Then our proposed contrastive learning strategy for radar points instance segmentation and its different settings are described in Section \ref{sec methods}. The information of the dataset, as well as the experimental results are presented in Section \ref{sec results}. Analyses are also made in this section. Finally, Section \ref{sec conclusion} summarizes our work and presents the future research direction.

\vspace{-1mm}
\section{RELATED WORK}\label{sec related}

There are mainly two types of point clouds: one is the dense point cloud such as the LiDAR point cloud, and the other is the sparse point cloud such as the radar point cloud.
In this section, we review related studies on instance segmentation with LiDAR point clouds and automotive radar detection points. Some of the methods are based on traditional algorithms, and others apply neural networks in a deep learning driven fashion. However, the former are restricted with their performance, and the latter need a lot of labeled training data. In addition, works on constrative learning are introduced, most of which focus on tasks about images.
These works only requires a few labeled data to train the models, and thus inspires us to propose our strategy.

\vspace{-1mm}
\subsection{Instance Segmentation with LiDAR Point Cloud}
There are many researches on LiDAR point cloud-based instance segmentation \cite{LiDAR_ins_seg_1}\cite{LiDAR_ins_seg_2}\cite{LiDAR_ins_seg_3}\cite{LiDAR_ins_seg_4}\cite{panoster}. Liu \emph{et al.} adopt the concept of LiDAR slices proposed in \cite{LiDAR_ins_seg_1_addition} and devises a Slice Growing algorithm to perform instance segmentation \cite{LiDAR_ins_seg_1}. The method first extracts and merges the major parts and then grows them by searching their neighbors. After that, semantic labels are predicted through an RNN for labeling move objects and a simple judgement algorithm for labeling static objects. However, this method is mostly based on traditional algorithm. Thus, as the environment of the ego-vehicle becomes more complicated, the performance of the method might be restricted. As a result, recent researches in the community shift the interest in deep learning-based methods.

Some researchers transform the LiDAR point cloud into a high-resolution Bird's Eye View (BEV) image which is then processed by a CNN-based model. Recent work includes \cite{LiDAR_ins_seg_2} which applies a revised stacked hourglass block to predict the semantic label and the center of the object for each foreground grid pixel, getting the predicted instances after grouping points and merging objects whose predicted centers are close enough. Xiong \emph{et al.} predict semantic segmentation results, center offsets and a contour map, and further predict a binary mask for the BEV image \cite{LiDAR_ins_seg_3}. After that, the final instance segmentation results are generated based on the predicted binary mask and the semantic information.

For deep learning methods taking raw LiDAR points as input, Behley \emph{et al.} treat the instance segmentation task as an object detection task, i.e., they generate a 3D bounding box for each instance \cite{LiDAR_ins_seg_4}. Gasperini \emph{et al.} propose a model called Panoster \cite{panoster}, which has a shared encoder and two independent decoders for semantic segmentation and instance segmentation, respectively. It is a completely learning-based method because its instance segmentation branch optimizes the loss based on the soft confusion matrix and do not require the external grouping process.

\vspace{-1mm}
\subsection{Instance Segmentation with Automotive Radar Detection Points} 

Due to the sparsity of automotive radar detection points, tasks such as instance segmentation with automotive radar detection points are more challenging than those with LiDAR point clouds. 
As the most commonly-used method for instance segmentation on radar points is clustering-based classification, some efforts are made by enhancing the clustering methods. For instance, Schumann \emph{et al.} modified the DBSCAN \cite{DBSCAN} algorithm and a score function is optimized in a traditional supervised machine learning strategy so that the parameters of DBSCAN could be automatically chosen to achieve a higher performance \cite{supervised_clustering}. However, each point cloud in training and testing is composed of all detection points within a certain time range, which may cause overlap, and the clustering parameters are still not related to the class of the instance, which restricts further improvements on performance.

Other work perform instance segmentation using deep learning methods. The work in \cite{radar_ins_seg_1} predicts the semantic label as well as a center shift vector for each detection point using gMLP \cite{gMLP} based PointNet++ \cite{PointNet++}, and applies DBSCAN \cite{DBSCAN} on the shifted points to obtain the instance information. The strategy in \cite{radar_ins_seg_2} is similar to the semantic segmentation-based clustering method in \cite{radar_ins_seg_1}, and a memory point cloud is introduced to %overcome the downside that the rich information among consecutive frames is not considered.
jointly consider the related information among consecutive frames to improve the segmentation performance. 
However, these deep learning-based methods require a large number of labeled training data, which restricts their application because of the difficulty to obtain enough point-wise labeled radar frames.

\vspace{-1mm}
\subsection{Contrastive Learning}
As a self-supervised learning strategy, contrastive learning which explores a proper representation of unlabeled data, plays a vital role when annotating training data is difficult. %There are usually two steps: representation learning and downstream task fine-tuning. 
To learn a good representation of the data, positive and negative samples need to be carefully chosen. Most of efforts are made in the field of computer vision, especially in image classification. For example, SimCLR \cite{SimCLR} uses data augmentation methods to generate a ``copy'' of each image. For every image, the ``copy'' is defined as its positive sample and other images as well as their augmentations are regarded as negative samples.

However, there are less literature about adopting contrastive learning for semantic segmentation, due to the difficulty of choosing positive and negative samples. The method proposed in \cite{contrast_for_semantic_3} inputs two different views of an image into two encoders, each with a dense projection head. The distances between each pair of pixels are calculated and the closest ones are selected as positive pairs.
Other researchers \cite{contrast_for_semantic_1}\cite{ contrast_for_semantic_2} use the pixel-wise label to define positive and negative samples. %in the consideration that even if few images are labeled, the pixels in those labeled images are sufficient for supervised contrastive learning. 
By doing that, in the case of a few labeled images in the dataset, the pixels in those labeled images are sufficient for supervised contrastive learning.
%To promote training, \cite{contrast_for_semantic_1, contrast_for_semantic_2} also propose different strategies: In order to get more negative samples, a memory bank which stores samples in the previous minibatch is introduced in \cite{contrast_for_semantic_1}; to make use of the unlabeled training data, a semi-supervised training setting is provided in \cite{contrast_for_semantic_2}.
To boost the performance, Chen \emph{et al.} introduce a memory bank to accumulate the negative samples in the used minibatch \cite{contrast_for_semantic_1}. Differently, Li \emph{et al.} use a semi-supervised training setting to make use of the unlabeled training data to enhance the outcome \cite{contrast_for_semantic_2}.

\vspace{-1mm}
\section{PROPOSED METHOD} \label{sec methods}

\begin{figure*}[b]
    \vspace{-4mm}
    \centering
    \includegraphics[scale=0.52]{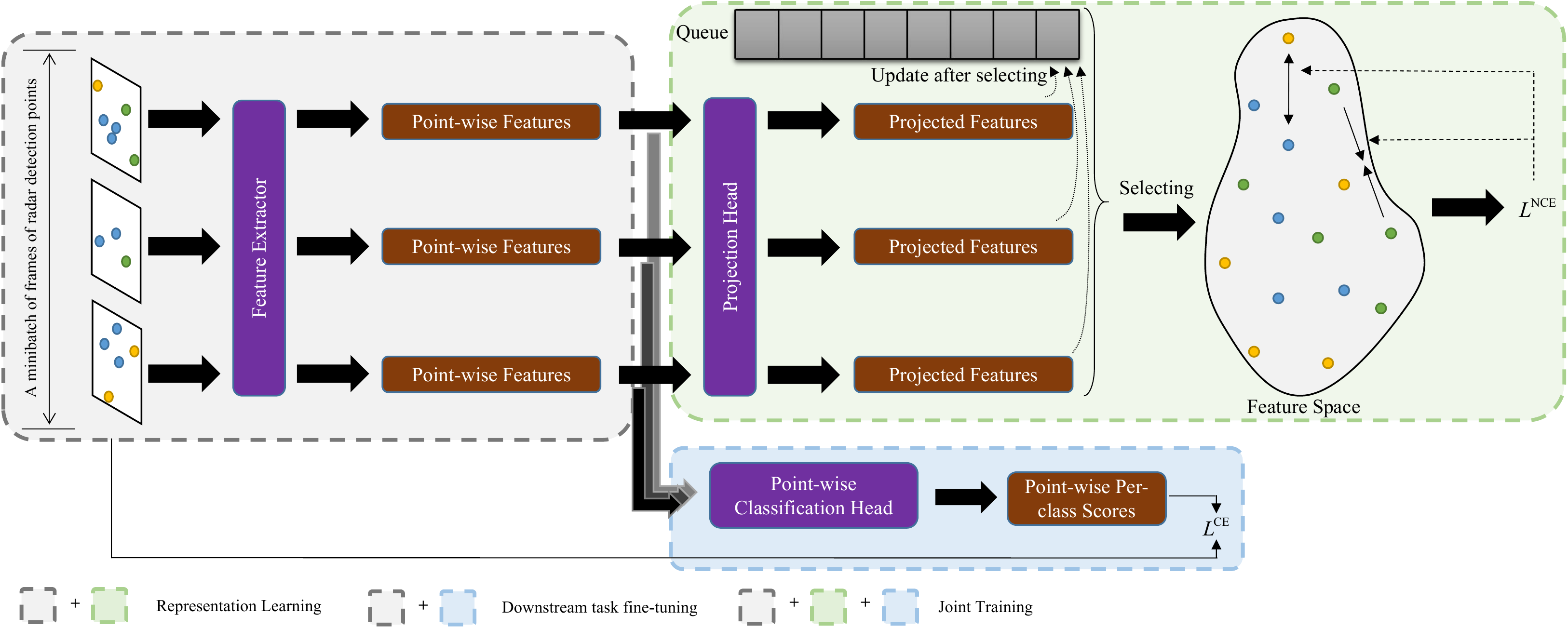}
    \vspace{-1mm}
    \caption{The training process of our proposed contrastive learning based model. 
    %$B$ is the batch size, $N_{\rm{sample}}$ represents the sample size and $n_{\rm{class}}$ denotes the number of classes. 
    The gray and green parts of the figure show the representation learning process, where features are projected by a projection head, points are selected from the frames in a minibatch as well as the queue, and the contrastive loss is calculated. The gray and blue parts show the downstream task fine-tuning process, during which the feature extractor is frozen, a point-wise classification head rather than the projection head is connected, and the cross entropy loss is calculated. The whole figure illustrates the joint training process. In this training strategy, both heads exist and the total loss is the weighted sum of the two losses.}
    \label{training process}
\end{figure*}

This section introduces the proposed instance segmentation method with the details of different settings in contrastive learning. In particular, our training can be performed in either fully-supervised setting or semi-supervised setting, which is similar to \cite{contrast_for_semantic_2}. The former uses the small amount of training data which have been labeled, while the latter takes both labeled and massive unlabeled training data as input.

\vspace{-1mm}
\subsection{Overview}
% Here write an Overview of Our Proposed Contrastive Learning Based Solution
%In this work, we adopt the semantic segmentation-based clustering method for the instance segmentation task. It has been demonstrated in \cite{radar_ins_seg_1} that semantic segmentation-based clustering is feasible in automotive mmWave radar real-time applications. 
In this work, we adopt the semantic segmentation-based clustering method for the instance segmentation task, as it has been demonstrated in \cite{radar_ins_seg_1} that this method is feasible in automotive mmWave radar real-time applications. 
%Moreover, the method proposed in \cite{radar_ins_seg_1} can achieve the state-of-the-art performance. 
To be specific, the points are input into a semantic segmentation model trained in a contrastive learning strategy, obtaining their predicted semantic label.
Then DBSCAN \cite{DBSCAN} is used to cluster the points belonging to the same class into different clusters, where each cluster represents an instance.

\begin{figure*}[b]
\vspace{-3mm}
\hrulefill
\begin{equation}
    \label{contrastive loss i}
    L^{\rm{NCE}}_i=\frac{1}{|P_i|}\sum_{i^+\in P_i}-\log\frac{\exp(f_i\cdot f_{i^+}/\tau)}{\exp(f_i\cdot f_{i^+}/\tau)+\sum_{i^-\in N_i}\exp(f_i\cdot f_{i^-}/\tau)}
\end{equation}
\end{figure*}
%由于公式会出现在添加位置的下一页顶部，所以将其放置在应该出现位置的上一页。

As mentioned in Section \ref{sec introduction}, %and section \ref{sec contrastive}, 
the process of contrastive learning consists of two steps: representation learning and downstream task fine-tuning. 
Representation learning is the critical step in the process because in this step, appropriate positive and negative samples are identified for the further training process.
%Representation learning is the major work, and defining appropriate positive and negative samples is of primary importance. 
In an image classification problem, data augmentation is widely used to help define positive and negative samples \cite{SimCLR}. However, data augmentation is challenging for radar detection points based semantic segmentation, and there is no suitable data augmentation method for \emph{radar} point cloud \emph{segmentation} tasks.
%There are a few reasons. Firstly, point-wise features are required for radar signals, rather than the global feature of a point cloud. 
%Secondly, there is no suitable data augmentation method for \emph{segmentation} tasks, particularly for \textcolor{red}{radar} point clouds.

Fortunately, many labeled points can be obtained even when the number of labeled frames of radar detection points is small. Thus, defining positive and negative samples according to the ground-truth label is feasible to train the model using a contrastive learning strategy. Furthermore, we can generate pseudo labels to make use of unlabeled training data, and define the positive and negative samples in a way similar to the semi-supervised setting in \cite{contrast_for_semantic_2}. More details are in the following subsections.

% \subsection{Baseline}
% Clustering + Random Forest Classifier for Defining Positive and Negative Samples
% 拟放在实验部分简要介绍

\subsection{Fully-Supervised Setting} \label{sec fully-supervised}

\subsubsection{Representation Learning}

The process of representation learning is illustrated in the gray and green parts of Fig. \ref{training process}. We define the positive and negative samples according to the ground-truth label in our fully-supervised setting due to sufficient labeled points, even if the number of labeled frames is small.
In detail, points with the same semantic label are defined as positive samples, and points with different semantic labels are negative samples. To balance the number of points in each class, the following operations are performed: in each minibatch, $n_{\rm{point}}$ points are randomly selected from $n_{\rm{minibatch}}$ frames, where the number of points in each class are the same; as a result, $N=n_{\rm{point}}/n_{\rm{class}}$ points for each class are selected, where $n_{\rm{class}}$ denotes the number of classes.
% 注：为了与输入网络时保证一个batch所有帧点数相同的而进行的采样（sample）区别，这里采用select表示在minibatch中采样。

To make sure that there is enough positive and negative samples (i.e., number of points in each class) in every minibatch, a queue is used to store the features of points in the previous minibatchs. That is, if there are not enough points in some classes, the remaining points will be selected from the queue, and after that the queue will be updated. Let $n_{i, j}^{\rm{minibatch}}$ denote the number of points belonging to the $j$-th class in the $i$-th minibatch, and $n_{i, j}^{\rm{queue}}$ denote the number of points belonging to the $j$-th class in the queue before the $i$-th minibatch. It should be noted that, while the maximum size of the queue is large enough, the situation that $n_{i, j}^{\rm{queue}} + n_{i, j}^{\rm{minibatch}} < N$ will not happen as long as the first minibatch satisfies $n_{1, j}^{\rm{minibatch}} > N$.

After getting $n_{\rm{point}}$ selections, the contrastive loss of each point in the minibatch is calculated, as shown in \eqref{contrastive loss i}, where $i$ is one of the $n_{\rm{point}}$ selected points and $\tau$ denotes the temperature parameter; $P_i$ and $N_i$ are the sets of positive and negative samples of point $i$, while $i^+$ and $i^-$ are an element of $P_i$ and $N_i$, respectively. In our case, $|P_i|=N-1$ and $|N_i|=(n_{\rm{class}}-1)\times N$. It is worth noting that a projection head is added to the backbone of our method, as shown in the green part of Fig. \ref{training process}, and $f_i$ in \eqref{contrastive loss i} represents the projected feature of point $i$.

The contrastive loss of the minibatch is the average loss of all selected points, i.e., 
\begin{equation}
\label{contrastive loss}
L^{\rm{NCE}}=\frac{1}{n_{\rm{point}}}\sum_{i\in S}L^{\rm{NCE}}_i,
\end{equation}
where $S$ is the set of selected points in the minibatch, which satisfies $|S|=n_{\rm{point}}$ and for each point $i$, $S=\{i\}\cup P_i\cup N_i$. The loss in \eqref{contrastive loss i} is also known as InfoNCE \cite{InfoNCE}, which is widely used in contrastive learning. By backpropagation, the features of the positive samples will be pulled together, and the features of the negative samples will be pushed away. Thus in the feature space, the features of negative samples can be dispersed, while the features of positive samples are gathered.

\subsubsection{Downstream Task Fine-Tuning}

The process of downstream task fine-tuning is illustrated in the gray and blue parts of Fig.~\ref{training process}. The representation learning step only makes the model learn a suitable feature representation method for the radar detection points-based semantic segmentation task. That is, the model outputs the extracted features rather than the predicted classes of the points. As a result, downstream task fine-tuning is applied to make the model further learn how to classify the points using the learned features.

During fine-tuning, the projection head is replaced by a point-wise classification head, as shown in Fig. \ref{training process}. Note that the variables in the backbone is frozen during the fine-tuning process and only the parameters in the newly connected head is adjustable. This training process is a typical transfer learning (using the labeled training data again) with cross entropy loss ($L^{\rm{CE}}$) for point-wise classification.

Finally, DBSCAN is employed on the classified points, and a grid search with the evaluating metric of mAP$_{0.5}$ is employed to find the best clustering parameters.

\vspace{-1mm}
\subsection{Semi-Supervised Setting}
In the fully-supervised setting, a large amount of unlabeled training data are not used. %Considering that more input data are likely to improve the performance, we need to find a proper way to utilize the unlabelled data.
% A fully supervised setting can only use the labeled data, which is typical in many image segmentation tasks. However, many points do not have a valid label in the segmentation problem for the sparse radar signal points. 
In order to use these unlabeled points to enhance the segmentation accuracy of the network, we incorporate the ``pseudo labels'' into the training process. Note that the ``pseudo labels'' are only generated for the radar detection points without a label.
%As no class information is available for the unlabeled training data, defining positive and negative samples like that in fully-supervised setting has a problem, so “pseudo labels” should be generated for the points of unlabeled data. 

A simple way to generate pseudo labels is to apply the model trained in fully-supervised setting to predict the class of each point, and regard it as the groundtruth label to help define the positive and negative samples. 
Note that the prediction is not 100\% accurate. If the accuracy is too low, the model may learn wrong representations for the points.
%However, the prediction is not 100\% correct, so the predicted class is called a pseudo label to distinguish from the groundtruth label. 
In practice, a confidence threshold $T\in(0,1)$ can be employed to increase the accuracy of the generated pseudo labels. The idea is that only prediction with a probability higher than the threshold will be retained for the training process.

The detailed training process of contrastive learning and downstream task fine-tuning in the semi-supervised setting is the same as that in the fully-supervised setting, as described in Section \ref{sec fully-supervised}.

\vspace{-1mm}
\subsection{Joint Training of Representation Learning and Downstream Task}
Inspired by \cite{contrast_for_semantic_1}, we merge the two steps of contrastive learning into a single overarching procedure. 
The process of joint training is illustrated in Fig. \ref{training process}. 
Note that it can be performed in either fully-supervised setting or semi-supervised setting. 
Specifically, the projection head and the point-wise classification head are connected to the feature extractor at the same time, and the total loss is calculated as
\begin{equation}
\label{total loss}
    L=L^{\rm{NCE}}+\alpha L^{\rm{CE}},
\end{equation}
where $\alpha>0$ is a weighting hyper-parameter. 

By adopting joint training instead of non-joint training, an improved performance can be obtained. More details are discussed in Section \ref{sec results}.

\begin{figure*}[t]
    \centering
    \includegraphics[scale=0.54]{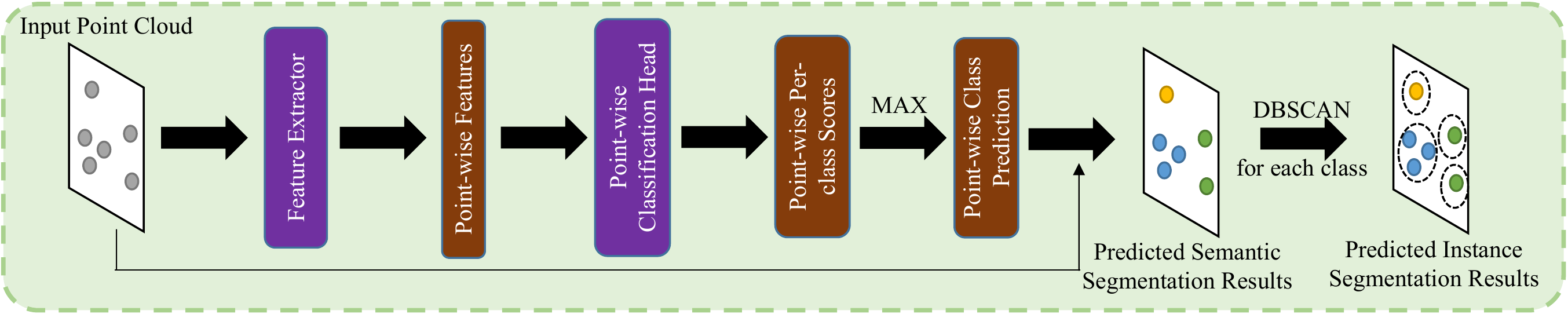}
    \vspace{-2mm}
    \caption{The inference process of our proposed contrastive learning based model. 
    %$N_{\rm{sample}}$ is the sample size and $n_{\rm{class}}$ denotes the number of classes. 
    During the inference process, only the point-wise classification head is connected, and DBSCAN is applied based on the semantic segmentation results to obtain the instance information.}
    \label{inference process}
    \vspace{-5mm}
\end{figure*}

\begin{figure*}[b]
    \centering
    \vspace{-5mm}
    \includegraphics[scale=0.54]{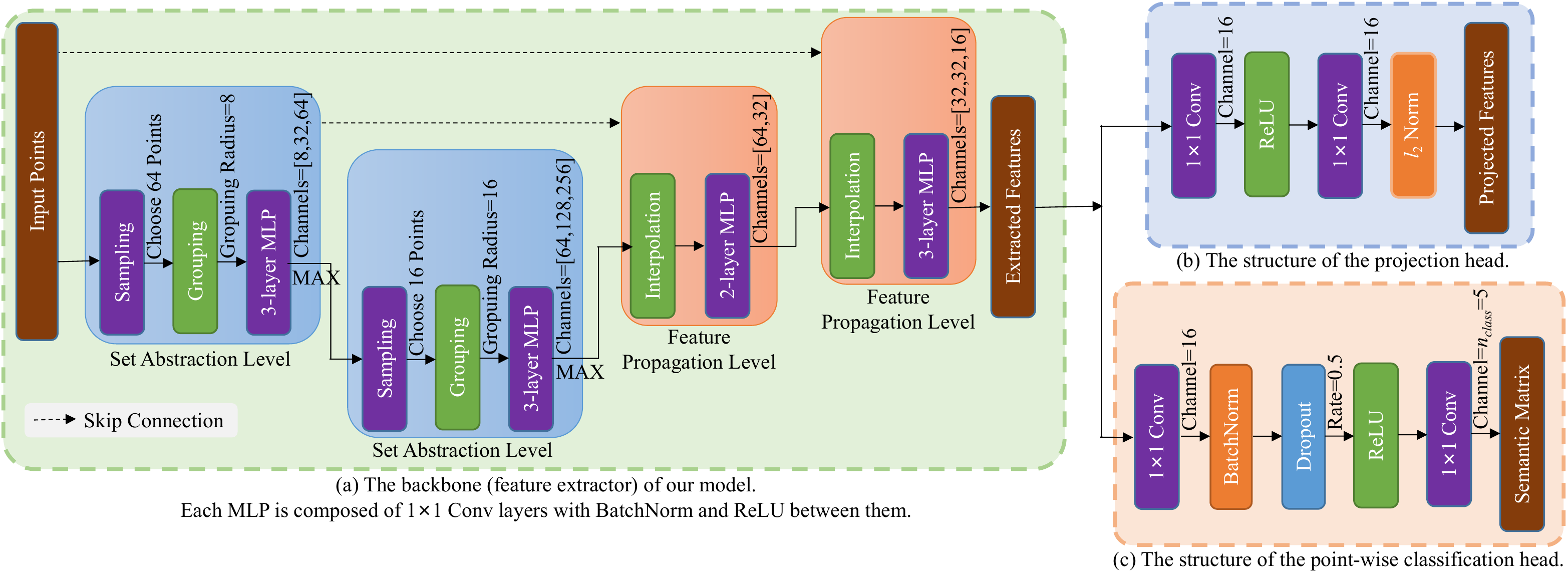}
    \caption{The detailed configuration of our proposed contrastive learning based model, including that of the feature extractor and the two heads. The structure and hyper-parameters of the feature extractor is the same as that in \cite{radar_ins_seg_1}.}
    \label{model}
\end{figure*}

\vspace{-1mm}
\subsection{Inference}
During the inference process, the projection head of the model, the queue for storing positive and negative samples and the minibatch selecting process are all discarded. As illustrated in Fig.~\ref{inference process}, the model takes the point cloud as input, where the feature extractor extracts point-wise features and the point-wise classification head predicts per-class score for each point. Finally, the DBSCAN is applied to obtain the predicted instance information.

\vspace{-1mm}
\section{EXPERIMENTS}\label{sec results}
In this section, a brief description of the dataset is presented first, and then our implementation details including the network architecture and hyper-parameters are introduced. After that, the experimental results are presented followed by some analyses and discussions. 

\vspace{-1mm}
\subsection{Dataset}
We choose RadarScenes \cite{radarscenes} as the dataset of our experiments. There are 158 sequences of data containing 1,556,684 frames of radar detection point clouds, and the coordinates, velocity, radar cross section (RCS) as well as other features of points are provided. 
%Only dynamic points are adopted as we only focus on the road users, and the 11 categories of non-static objects are merged into 5 (i.e., car, pedestrian, group of pedestrians, large vehicle and two-wheeler; as a result, $n_{\rm{class}}=5$) \textcolor{red}{due to lack of data in some classes}.
Only dynamic points are adopted as we only focus on the road users, and the 11 categories of non-static objects are merged into 5 (i.e., car, pedestrian, group of pedestrians, large vehicle and two-wheeler; as a result, $n_{\rm{class}}=5$) due to lack of data in some classes. These two settings are suggested by the guidance of the RadarScenes dataset.

In our experiments, the dataset is split into 8:1:1, which is the ratio of frames in the training set, validation set and test set. Furthermore, we divide the training set into different proportions of labeled data and unlabeled data, whose labels are discarded in our experiments to simulate the situation that most of the data do not have any annotation.

\begin{table*}[b]
\centering
\vspace{-4mm}
\begin{threeparttable}[b]
\caption{Results of Different Strategies for Instance Segmentation on Test Data}
\label{results}
\begin{tabular}{|cccc||c|c|}
\hline
\multicolumn{4}{|c||}{Learning Strategy}  & mCov(\%) & mAP$_{0.5}$(\%) \\ \hline
\multicolumn{1}{|c|}{\multirow{4}{*}{\begin{tabular}[c]{@{}c@{}}Supervised\\Learning\end{tabular}}}    & \multicolumn{1}{c||}{5\% Training Data}                  & \multicolumn{2}{c||}{\multirow{2}{*}{Clustering + Random Forest Classifier Classification}}                                                 & 73.18    & 68.19   \\
\multicolumn{1}{|c|}{}                                        & \multicolumn{1}{c||}{100\% Training Data}                & \multicolumn{2}{c||}{}                                                                                                                      & 79.54\tnote{*}    & 76.09\tnote{*}   \\ \cline{2-6} 
\multicolumn{1}{|c|}{}                                        & \multicolumn{1}{c||}{5\% Training Data}                  & \multicolumn{2}{c||}{\multirow{2}{*}{Semantic Segmentation + Clustering}}                                                                   & 75.59    & 70.86   \\
\multicolumn{1}{|c|}{}                                        & \multicolumn{1}{c||}{100\% Training Data}                & \multicolumn{2}{c||}{}                                                                                                                      & 82.21\tnote{*}    & 77.96\tnote{*}   \\ \hline
\multicolumn{1}{|c|}{\multirow{13}{*}{\begin{tabular}[c]{@{}c@{}}Contrastive\\Learning\end{tabular}}} & \multicolumn{1}{c||}{5\% Labeled Data}                   & \multicolumn{2}{c||}{Baseline: Clustering + Classification for Defining Positive and Negative Samples}                                      & 71.92    & 66.23   \\ \cline{2-6} 
\multicolumn{1}{|c|}{}                                        & \multicolumn{1}{c||}{\multirow{4}{*}{5\% Labeled Data}}  & \multicolumn{1}{c|}{\multirow{2}{*}{\begin{tabular}[c]{@{}c@{}}Non-joint Training (Representation\\Learning + Downstream Task Fine-tuning)\end{tabular}}} & Fully-supervised Setting & 77.47    & 72.41   \\
\multicolumn{1}{|c|}{}                                        & \multicolumn{1}{c||}{}                                   & \multicolumn{1}{c|}{}                                                                                           & Semi-supervised Setting  & 78.58    & 73.82   \\ \cline{3-6} 
\multicolumn{1}{|c|}{}                                        & \multicolumn{1}{c||}{}                                   & \multicolumn{1}{c|}{\multirow{2}{*}{Joint Training}}                                                            & Fully-supervised Setting & 77.57    & 72.83   \\
\multicolumn{1}{|c|}{}                                        & \multicolumn{1}{c||}{}                                   & \multicolumn{1}{c|}{}                                                                                           & Semi-supervised Setting  & 79.00    & 74.01   \\ \cline{2-6} 
\multicolumn{1}{|c|}{}                                        & \multicolumn{1}{c||}{\multirow{3}{*}{10\% Labeled Data}} & \multicolumn{1}{c|}{\multirow{2}{*}{\begin{tabular}[c]{@{}c@{}}Non-joint Training (Representation\\Learning + Downstream Task Fine-tuning)\end{tabular}}} & Fully-supervised Setting & 79.43    & 74.83   \\
\multicolumn{1}{|c|}{}                                        & \multicolumn{1}{c||}{}                                   & \multicolumn{1}{c|}{}                                                                                           & Semi-supervised Setting  & 80.03    & 75.61   \\ \cline{3-6} 
\multicolumn{1}{|c|}{}                                        & \multicolumn{1}{c||}{}                                   & \multicolumn{1}{c|}{Joint Training}                                                                             & Fully-supervised Setting & 79.59    & 75.06   \\ \cline{2-6} 
\multicolumn{1}{|c|}{}                                        & \multicolumn{1}{c||}{\multirow{2}{*}{20\% Labeled Data}} & \multicolumn{1}{c|}{\multirow{4}{*}{\begin{tabular}[c]{@{}c@{}}Non-joint Training (Representation\\Learning + Downstream Task Fine-tuning)\end{tabular}}} & Fully-supervised Setting & 80.72    & 76.48   \\
\multicolumn{1}{|c|}{}                                        & \multicolumn{1}{c||}{}                                   & \multicolumn{1}{c|}{}                                                                                           & Semi-supervised Setting  & 81.06    & 76.82   \\ \cline{2-2} \cline{4-6} 
\multicolumn{1}{|c|}{}                                        & \multicolumn{1}{c||}{40\% Labeled Data} & \multicolumn{1}{c|}{}                                                                                           & Fully-supervised Setting & 81.76    & 77.66   \\ \cline{2-2} \cline{4-6} 
\multicolumn{1}{|c|}{}                                        & \multicolumn{1}{c||}{100\% Labeled Data}                 & \multicolumn{1}{c|}{}                                                                                           & Fully-supervised Setting & \textbf{82.73}    & \textbf{79.01}   \\ \hline
\end{tabular}
\begin{tablenotes}
    \item[*] The results are from \cite{radar_ins_seg_1}.
%    \item[1] The ``5\% training data'' means to use the same training data as those of fully-supervised setting in contrastive learning, which account for 5\% of the whole training set. 
%    \item[2] The ``100\% training data'' means to use entire training set and the labels of all training data are kept in this setting. The experimental results are from \cite{radar_ins_seg_1}.
\end{tablenotes}
\end{threeparttable}
\end{table*}

\vspace{-1mm}
\subsection{Experiment Configurations}
In this study, we selected PointNet++ \cite{PointNet++} as the backbone of our model. The model structure is the same as that in \cite{radar_ins_seg_1}, which is shown in Fig. \ref{model}(a). The projection head used in representation learning is a two-layer MLP with ReLU and $l_2$ normalization is applied to the output of the MLP, whose structure is illustrated in Fig. \ref{model}(b); The point-wise classification head is another two-layer MLP with Batch Normalization, ReLU and Dropout, whose structure is shown in Fig. \ref{model}(c).

PointNet++ can only take a batch of point clouds with the same number of points, so we set the sample size ($N_{\rm{sample}}$) to 100 after calculating some statistics of the frames, i.e., 100 points are sampled from all points in each frame; in most case, there are not enough points in a frame, so some points may be sampled more than once. During representation learning, the duplicated points will be removed before selecting points in a minibatch, where the number of selected points ($n_{\rm{point}}$) is set as 250. Therefore, $N=50$ points of each class are selected in a minibatch.

The scheduler of cosine annealing warm restarts and the ADAM optimizer is adopted during training. The initial learning rate is 1e-2 in representation learning and joint training, while it is set to 5e-4 during downstream task fine-tuning. In fully-supervised setting, the batch size is 512, and the minibatch size ($n_{\rm{minibatch}}$) is set as 32; in semi-supervised setting, the batch and minibatch sizes are both doubled.

\vspace{-1mm}
\subsection{Experimental Results and Analysis}
Following \cite{radar_ins_seg_1}, we choose 
the mean coverage (mCov) and mean average precision with the IoU threshold of 0.5 (mAP$_{0.5}$) as the evaluation metrics of our experiments for fair comparison. Table \ref{results} exhibits the results on test data with different training strategies and methods, including the clustering-based classification method and %supervised training strategy (
semantic segmentation-based clustering method %)
proposed in \cite{radar_ins_seg_1}. 
The former one is the most popular method in the industry and the latter achieves state-of-the-arts performance, so we employ them as our baselines of supervised learning. Correspondingly, for the baseline method of contrastive learning, we apply this idea to generating pseudo labels and defining the positive and negative samples.

Specifically, the point cloud is clustered by DBSCAN first and a random forest classifier trained on the labeled training data is applied on unlabeled training data to generate pseudo labels, according to which the positive and negative samples are defined. After that the whole training dataset is sent to the model and contrastive learning is performed, and the details are the same as that described in Section \ref{sec fully-supervised}. The reason of the poor performance is the high inaccuracy in the generated pseudo labels, which proves the importance of defining appropriate positive and negative samples. For contrastive learning with different proportions of labeled training data, some of the experiments are omitted after we found the trend of the results obvious.

By comparing the results of different training strategies, we can draw the following three conclusions from Table \ref{results}:
\begin{itemize}
\item \textbf{Contrastive learning performs better than supervised learning in the condition that few labeled data are provided.} The model trained in contrastive learning (non-joint training in fully-supervised setting) strategy attains around 2\% improvement on performance compared to that trained in supervised learning when 5\% of the training data are used. By contrasting the features of positive and negative samples, the model learns to extract more appropriate features which are helpful to the later semantic segmentation task, but traditional supervised learning suffers from the lack of the training data.
\item \textbf{Contrastive learning in semi-supervised setting outperforms that in fully-supervised setting.} 
%The 1.5\% gain after changing the fully-supervised setting to semi-supervised setting in both non-joint training and joint training
When using 5\% labeled training data, there exists an 1.5\% gain after changing the fully-supervised setting to semi-supervised setting in both non-joint training and joint training. The phenomenon shows an evidence that more training data is beneficial to contrastive learning even if their labels are not completely correct.
In addition, as the proportion of labeled training data increases, the performance gain decreases because there are less unlabeled training data. This inspires us that we could get performance improvement by collecting more frames of radar points even though they are unlabeled.
\item \textbf{Joint training performs better than non-joint training.} By merging instead of separating the two steps of contrastive learning, there is a little improvement (0.2\%-0.4\%) on performance, which may be caused by the close relation between the representation learning and downstream task learning. The downstream task learning makes the feature extractor learn to extract features that are easy to be classified. For example, the features of points in different classes lie in different regions of the feature space, which is similar to the purpose of representation learning. The representation learning whose aim is to make the extracted features distributed like clusters in the feature space also makes the points easy to be classified. As a result, the two tasks in joint training can help each other so that the training is facilitated.
\end{itemize}

%It can be seen from Table \ref{results} that the performance of supervised learning with 100\% training data is still higher than that of our method. Even so, the experimental results of our methods are acceptable because only 1/20 labeled training data are available in these settings.
It can be seen from Table \ref{results} that the performance of our method with 40\% labeled training data is close to that of supervised learning with 100\% training data. Moreover, when all training data are labeled, our method outperforms the supervised learning strategy proposed in \cite{radar_ins_seg_1} by about 0.5\% mCov and 1\% mAP$_{0.5}$, showing the superiority of our strategy.

\begin{figure}[t]
    \centering
    \includegraphics[scale=1]{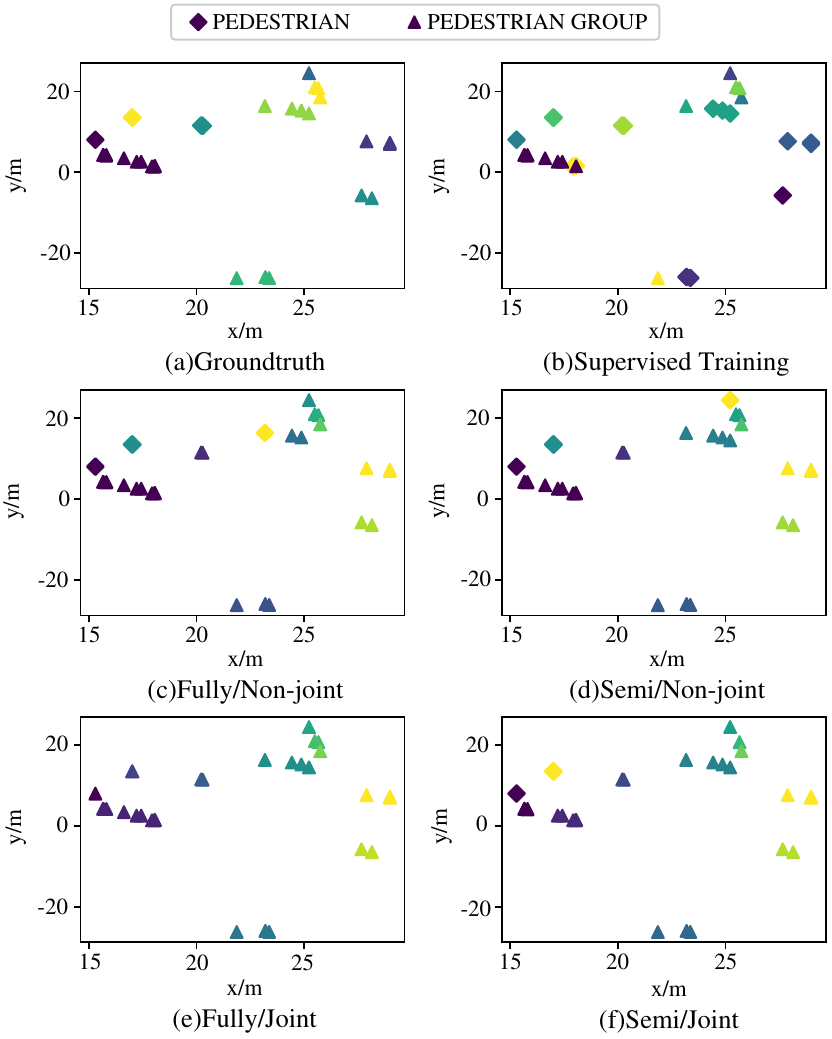}
    \vspace{-3mm}
    \caption{The instance segmentation results of an example frame. 5\% labeled training data are used to train the models. 
    As the sampling points are different during the inference process of each model, some points are missing in some subfigures. 
    Different markers denotes different classes, while different colors in the same class represents different instances in this class.}
    \label{example frame}
    \vspace{-5mm}
\end{figure}

The instance segmentation results of an example frame is shown in Fig.~\ref{example frame}. It is obvious that the model trained in joint/semi-supervised contrastive learning achieves the highest accuracy, while the model trained in supervised learning makes the most mistakes in the example frame.

Although the training time of different training strategies of contrastive learning in Table \ref{results} varies greatly, all the models have similar inference time (about 25ms per frame on an Intel Core i5-6300HQ CPU and with 8GB RAM) because the structure of feature extractor and the point-wise classification head is the same. Also, the storage memory of the models, as well as the number of parameters in the models is the same, which is 331KB and 75.245K respectively.

\vspace{-1mm}
\section{CONCLUSION} \label{sec conclusion}
In this work, we propose a contrastive learning strategy for radar detection points-based instance segmentation, to address the challenge of insufficient point-wise annotation of radar detection points.
%As the first method which employs a contrastive learning strategy for radar detection points-based instance segmentation, our work can automatically generate pseudo labels to improve the neural network's performance, to address the insufficient point-wise annotation of radar detection points. 
We experimented that four different training settings, including fully-supervised setting/non-joint training, semi-supervised setting/non-joint training, fully-supervised setting/joint training, and semi-supervised setting/joint training, can be applied on the proposed contrastive learning strategy while keeping a small model size and low inference time. This result shows that it is feasible to embed the proposed model on an automotive radar-based ADAS product. 
%Four different training settings, including fully-supervised setting/non-joint training, semi-supervised setting/non-joint training, fully-supervised setting/joint training and semi-supervised setting/joint training, can be applied on the strategy while keeping the inference time low and model size small. Thus, it is feasible to embed the model on a automotive radar based ADAS product.

%The performance of our strategy with a few labeled training data is comparable to that of the supervised learning strategy when a large number of labeled training data are provided. If the number of labeled training data is reduced to 5\% which is the same as that used in our contrastive learning strategy, the performance of the supervised learning strategy becomes lower than that of ours, proving the superiority of our method.

Further, we demonstrated via experiments that the performance of our proposed contrastive learning strategy with a few labeled training data is comparable to that of the supervised learning strategy with a large number of labeled training data. Moreover, when only a small amount of labeled training data is available for both supervised learning and our contrastive learning strategy, the performance of the supervised learning strategy becomes lower than that of ours, proving the superiority of our method.

However, our training strategy requires a longer training time compared to supervised learning. Our following work will focus on further improving the performance of proposed strategy while aiming to reduce the required training time.

%The disadvantage of our proposed approach is that the training time is too long, so that future research will be carried on to shorten the training time. Furthermore, we find that a little more labeled training data result in huge improvement in performance, more investigations will be executed to understand how much the performance can be improved by using 100\% labeled data with our contrastive learning strategy.

%Future research will be carried on to investigate more efficient approaches on defining the positive and negative samples for radar detection points based contrastive learning strategy, e.g., defining the positive and negative samples over consecutive radar frames by assisting of multi-object tracking or certain data association.

\addtolength{\textheight}{-15.8cm}
% This command serves to balance the column lengths on the last page of the document manually. It shortens the textheight of the last page by a suitable amount.  This command does not take effect until the next page so it should come on the page before the last. Make sure that you do not shorten the textheight too much.

%%%%%%%%%%%%%%%%%%%%%%%%%%%%%%%%%%%%%%%%%%%%%%%%%%%%%%%%%%%%%%%%%%%%%%%%%%%%%%%%
%\section*{APPENDIX}

%Appendixes should appear before the acknowledgment.

%\section*{ACKNOWLEDGMENT}

%The preferred spelling of the word ÒacknowledgmentÓ in America is without an ÒeÓ after the ÒgÓ. Avoid the stilted expression, ÒOne of us (R. B. G.) thanks . . .Ó  Instead, try ÒR. B. G. thanksÓ. Put sponsor acknowledgments in the unnumbered footnote on the first page.

%%%%%%%%%%%%%%%%%%%%%%%%%%%%%%%%%%%%%%%%%%%%%%%%%%%%%%%%%%%%%%%%%%%%%%%%%%%%%%%%

%References are important to the reader; therefore, each citation must be complete and correct. If at all possible, references should be commonly available publications.

\vspace{-1mm}
%\bibliographystyle{IEEEtran}
%\bibliography{reference}
%\begin{thebibliography}{99}

\end{document}